\documentclass[runningheads]{llncs}

\usepackage{graphicx}
\usepackage{subfiles}
\usepackage{comment}
\usepackage{amsmath,amssymb} 
\usepackage{amsfonts}
\usepackage{color, soul}
\usepackage{color, colortbl}
\usepackage{multirow}
\usepackage{makecell}
\usepackage{xcolor}
\usepackage{array}
\usepackage{booktabs}
\usepackage{hyperref}

\definecolor{LightCyan}{rgb}{0.88,1,1}

\newcolumntype{L}{>{\centering\arraybackslash}m{0.1\textwidth}}
\newcolumntype{K}{>{\centering\arraybackslash}m{0.15\textwidth}}

\definecolor{bluegray}{RGB}{102, 153, 204}
\definecolor{steelblue}{RGB}{70, 130, 180}

\newcommand{\ra}[1]{\renewcommand{\arraystretch}{#1}}

\newif\ifdraft
\drafttrue

\definecolor{orange}{rgb}{1,0.5,0}
\definecolor{pink}{rgb}{0.98, 0.38, 0.5}
\definecolor{darkgreen}{rgb}{0.055, 0.490, 0.016} 

\ifdraft
 \usepackage[normalem]{ulem}
 \newcommand{\RS}[1]{{\color{red}{\bf RS: #1}}}
 
 \newcommand{\PMN}[1]{{\color{orange}{\bf PMN: #1}}}
 
 \newcommand{\STM}[1]{{\color{blue}{\bf STM: #1}}}

\else
 \newcommand{\sout}[1]{}
 \newcommand{\RS}[1]{{\color{red}{}}}
 
 \newcommand{\PMN}[1]{{\color{red}{}}}
 
 \newcommand{\STM}[1]{{\color{red}{}}}
 
  \newcommand{\ARGUMENT}[1]{{\color{gray}{}}}
 
\fi

\newcommand{\real}{\mathbb{R}}

\newcommand{\x}{\mathbf{x}}

\newcommand{\q}{\mathbf{q}}
\newcommand{\g}{\mathbf{g}}

\newcommand{\m}{\mathbf{m}}

\newcommand{\W}{\mathbf{W}}

\newcommand{\att}{\textrm{att}}

\newcommand{\eg}{\emph{e.g.}}
\newcommand{\ie}{\emph{i.e.}}

\newcommand{\ours}{\textbf{Ours}}

\DeclareMathOperator*{\argmax}{arg\,max}

\begin{document}

\title{Localized Questions in \\ Medical Visual Question Answering}


\titlerunning{Localized Questions in  Medical Visual Question Answering}


\author{Sergio Tascon-Morales, Pablo Márquez-Neila, Raphael Sznitman}

\authorrunning{Tascon-Morales et al.}

\institute{University of Bern, Bern, Switzerland\\ \email{\{sergio.tasconmorales, pablo.marquez, raphael.sznitman\}@unibe.ch}}

\maketitle          

\begin{abstract}
Visual Question Answering (VQA) models aim to answer natural language questions about given images. Due to its ability to ask questions that differ from those used when training the model, medical VQA has received substantial attention in recent years. However, existing medical VQA models typically focus on answering questions that refer to an entire image rather than where the relevant content may be located in the image. Consequently, VQA models are limited in their interpretability power and the possibility to probe the model about specific image regions. This paper proposes a novel approach for medical VQA that addresses this limitation by developing a model that can answer questions about image regions while considering the context necessary to answer the questions. Our experimental results demonstrate the effectiveness of our proposed model, outperforming existing methods on three datasets. 


\end{abstract}
\section{Introduction}
\label{sec:intro}

Visual Question Answering (VQA) models are neural networks that answer natural language questions about an image~\cite{antol2015vqa,goyal2017making,hudson2019gqa,tan2019lxmert}. The capability of VQA models to interpret natural language questions is of great appeal, as the range of possible questions that can be asked is vast and can differ from those used to train the models. This has led to many proposed VQA models for medical applications in recent years~\cite{ImageCLEFVQA_Med2018,liu2019effective,liao2020aiml,vu2020question,zhan2020medical,gong2021cross,yu2023question}. These models can enable clinicians to probe the model with nuanced questions, thus helping to build confidence in its predictions.

\begin{figure}[!t]
\begin{center}
\includegraphics[width=0.9\textwidth]{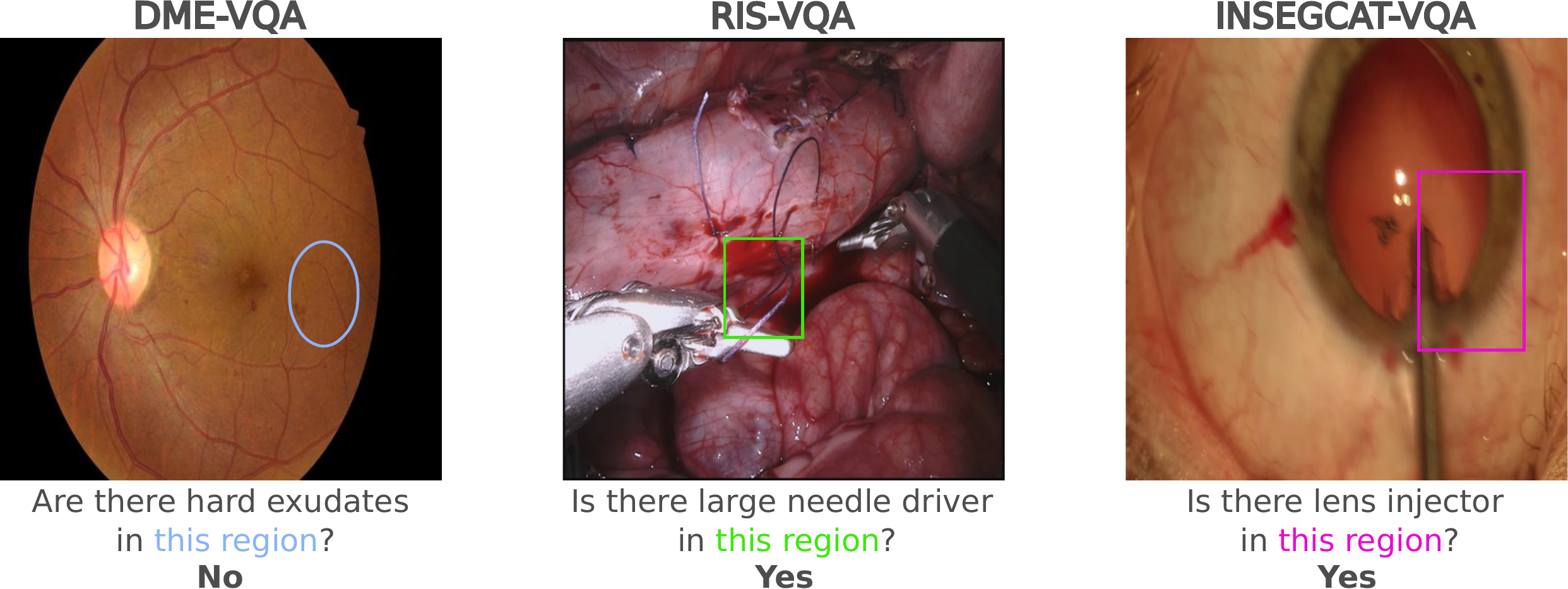}
\caption{Examples of localized questions. In some cases (RIS-VQA and INSEGCAT-VQA), the object mentioned in the question is only partially present in the region. We hypothesize that context can play an important role in answering such questions.}
\label{fig:examples_data}
\end{center}
\end{figure}

Recent work on medical VQA has primarily focused on building more effective model architectures~\cite{gong2021cross,ren2020cgmvqa,vu2020question} or developing strategies to overcome limitations in medical VQA datasets~\cite{Nguyen19,liu2021slake,pelka2018radiology,do2021multiple,vu2020question}. Another emerging trend is to enhance VQA performance by addressing the consistency of answers produced~\cite{tascon2022consistency}, particularly when considering entailment questions (\ie, the answer to ``Is the image that of a healthy subject?" should be consistent with the answer to ``Is there a fracture in the tibia?"). Despite these recent advances, however, most VQA models restrict to questions that consider the entire image at a time. Specifically, VQA typically uses questions that address content within an image without specifying where this content may or may not be in the image. Yet the ability to ask specific questions about regions or locations of the image would be highly beneficial to any user as it would allow fine-grained questions and model probing. For instance, Fig.~\ref{fig:examples_data} illustrates examples of such \emph{localized questions} that combine content and spatial specifications. 

To this day, few works have addressed the ability to include location information in VQA models. In~\cite{mani2020point}, localization information is posed in questions by constraining the spatial extent to a point within bounding boxes yielded by an object detector. The model then focuses its attention on objects close to this point. However, the method was developed for natural images and relies heavily on the object detector to limit the attention extent, making it difficult to scale in medical imaging applications. Alternatively, the approach from~\cite{vu2020question} answers questions about a pre-defined coarse grid of regions by directly including region information into the question (\eg, ``Is grasper in (0,0) to~(32,32)?"). This method relies on the ability of the model to learn a spatial mapping of the image and limits the regions to be on a fixed grid. Localized questions were also considered in~\cite{tascon2022consistency}, but the region of interest was cropped before being presented to the model, assuming that the surrounding context is irrelevant for answering this type of question.

To overcome these limitations, we propose a novel VQA architecture that alleviates the mentioned issues. At its core, we hypothesize that by allowing the VQA model to access the entire images and properly encoding the region of interest, this model can be more effective at answering questions about regions. To achieve this, we propose using a multi-glimpse attention mechanism~\cite{ben2017mutan,vu2020question,tascon2022consistency} restricting its focus range to the region in question, but only after the model has considered the entire image. By doing so, we preserve contextual information about the question and its region. We evaluate the effectiveness of our approach by conducting extensive experiments on three datasets and comparing our method to state-of-the-art baselines. Our results demonstrate performance improvements across all datasets. 

\section{Method}
\label{sec:method}

\begin{figure}[!t]
\begin{center}
\includegraphics[width=0.99\textwidth]{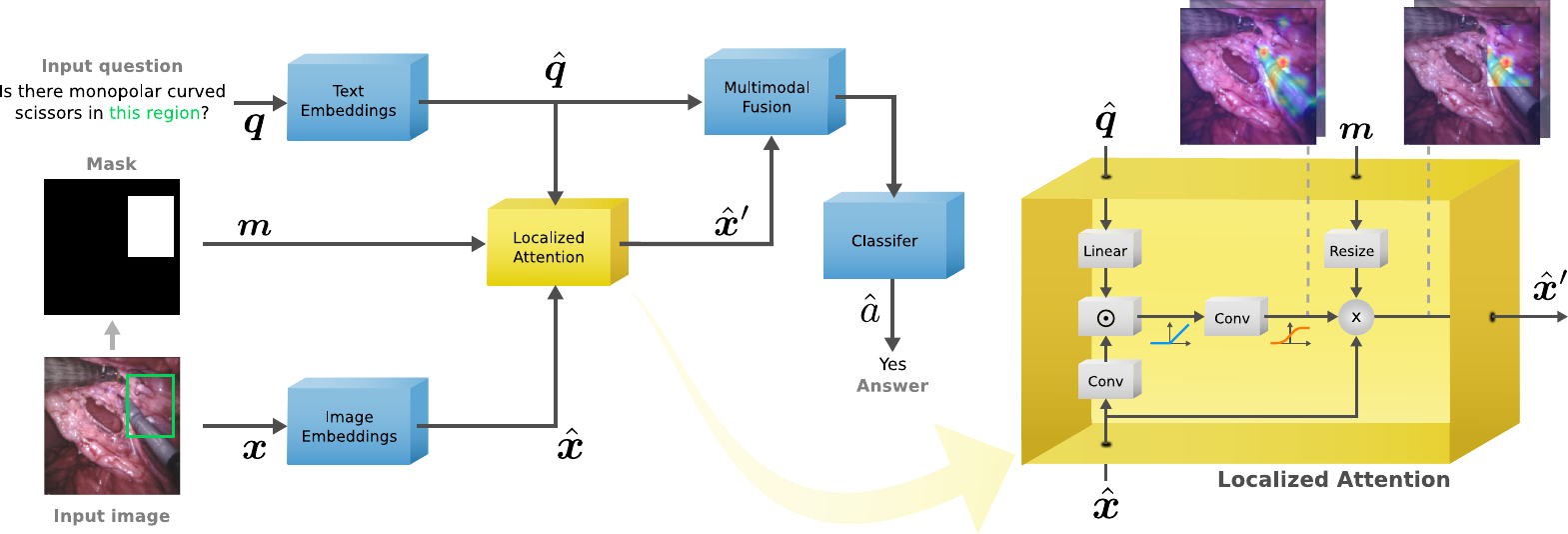}
\caption{\textbf{Left:} Proposed VQA architecture for localized questions. The Localized Attention module allows the region information to be integrated into the VQA while considering the context necessary to answer the question. \textbf{Right:} Localized Attention module.  
}
\label{fig:method}
\end{center}
\end{figure}

Our method extends a VQA model to answer localized questions. We define a \emph{localized question} for an image~$\x$ as a tuple~$(\q, \m)$, where $\q$~is a question, and $\m$~is a binary mask of the same size as~$\x$ that identifies the region to which the question pertains. Our VQA model~$p_\theta$, depicted in Fig.~\ref{fig:method}, accepts an image and a localized question as input and produces a probability distribution over a finite set~$\mathcal{A}$ of possible answers. The final answer of the model~$\hat{a}$ is the element with the highest probability,
\begin{equation}
    \hat{a} = \argmax_{a\in\mathcal{A}} p_\theta(a\mid \q, \x, \m).
\end{equation}
The model proceeds in three stages to produce its prediction: input embedding, localized attention, and final classification.

\textbf{Input embedding.} The question~$\q$ is first processed by an LSTM~\cite{hochreiter1997long} to produce an embedding~$\hat{\q}\in\real^Q$. Similarly, the image~$\x$ is processed by a ResNet-152~\cite{he2016deep} to produce the feature map~$\hat{\x}\in\real^{C\times{}H\times{}W}$.

\textbf{Localized attention.}
An attention mechanism uses the embedding to determine relevant parts of the image to answer the corresponding question. Unlike previous attention methods, we include the region information that the mask defines. Our \emph{localized attention} module (Fig.~\ref{fig:method} right) uses both descriptors and the mask to produce multiple weighted versions of the image feature map,~$\hat{\x}'=\att(\hat{\q}, \hat{\x}, \m)$. To do so, the module first computes an attention map~$\g\in\real^{G\times{}H\times{}W}$ with $G$~glimpses by applying unmasked attention~\cite{kim2016hadamard,vu2020question} to the image feature map and the text descriptor. The value of the attention map at location~$(h, w)$ is computed as,
\begin{equation}
    \g_{:hw} = \textrm{softmax}\left(\W^{(g)}\cdot\textrm{ReLU}\left(\W^{(x)}\hat{\x}_{:hw} \odot \W^{(q)}\hat{\q}\right)\right),
\end{equation}
where the index ${:}hw$ indicates the feature vector at location~$(h, w)$, $\W^{(x)}\in\real^{C'\times C}$, $\W^{(q)}\in\real^{C'\times Q}$, and $\W^{(g)}\in\real^{G\times C'}$ are learnable parameters of linear transformations, and $\odot$~is the element-wise product. In practice, the transformations $\W^{(x)}$ and $\W^{(g)}$ are implemented with $1\times{}1$~convolutions and all linear transformations include a dropout layer applied to its input. The image feature maps~$\hat{\x}$ are then weighted with the attention map and masked with~$\m$ as,
\begin{equation}
    \hat{\x}'_{cghw} = \g_{ghw} \cdot \hat{\x}_{chw} \cdot \left(\m\downarrow_{H\times{}W}\right)_{hw},
\end{equation}
where $c$ and~$g$ are the indexes over feature channels and glimpses, respectively, $(h, w)$~is the index over the spatial dimensions, and $\m\downarrow_{H\times{}W}$~denotes a binary downsampled version of~$\m$ with the spatial size of~$\hat{\x}$. This design allows the localized attention module to compute the attention maps using the full information available in the image, thereby incorporating context into them before being masked to constrain the answer to the specified region.

\textbf{Classification.} The question descriptor~$\hat{\q}$ and the weighted feature maps~$\hat{\x}'$ from the localized attention are vectorized and concatenated into a single vector of size~$C\cdot{}G + Q$ and then processed by a multi-layer perceptron classifier to produce the final probabilities.

\textbf{Training.} The training procedure minimizes the standard cross-entropy loss over the training set updating the parameters of the LSTM encoder, localized attention module, and the final classifier. The training set consists of triplets of images, localized questions, and the corresponding ground-truth answers. As in~\cite{antol2015vqa}, the ResNet weights are fixed with pre-trained values, and the LSTM weights are updated during training.

\section{Experiments and results}
\label{sec:experiments_and_results}

We compare our model to several baselines across three datasets and report quantitative and qualitative results. Additional results are available in the supplementary material. 

\subsection{Datasets}
\label{subsec:datasets}
\begin{figure}[t]
\begin{center}
\includegraphics[width=\textwidth]{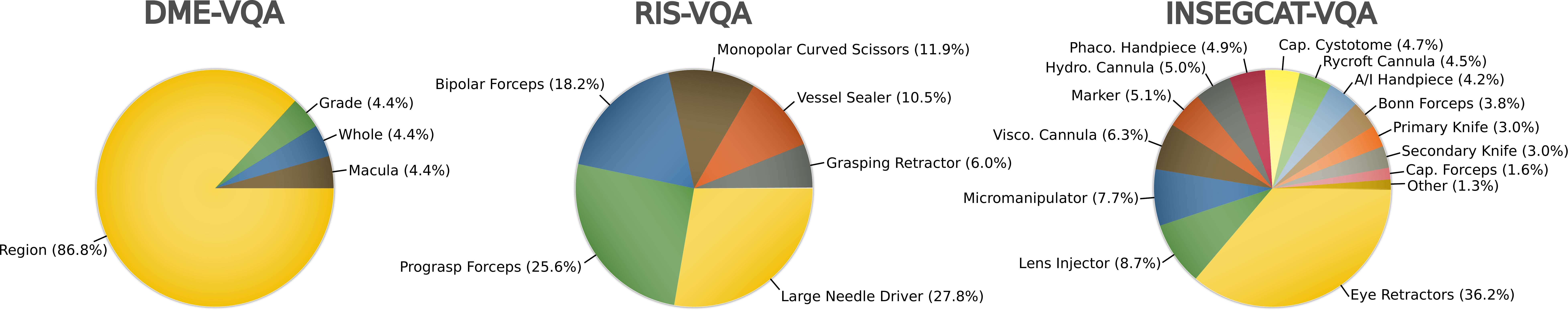}
\caption{Distribution by question type (DME-VQA) and by question object (RIS-VQA and INSEGCAT-VQA).}
\label{fig:object_distribution}
\end{center}
\end{figure}

We evaluate our method on three datasets containing questions about regions which we detail here. The first dataset consists of an existing retinal fundus VQA dataset with questions about the image's regions and the entire image. The second and third datasets are generated from public segmentation datasets but  use the method described in~\cite{vu2020question} to generate a VQA version with region questions. 
\begin{description}
    \item[DME-VQA~\cite{tascon2022consistency}.] 679~fundus images containing questions about entire images (\eg,~``what is the DME risk grade?") and about randomly generated circular regions (\eg,~``are there hard exudates in this region?"). The dataset comprises 9'779~question-answer~(QA) pairs for training, 2'380~for validation, and 1'311 for testing.
    \item[RIS-VQA.] Images from the 2017~Robotic Instrument Segmentation dataset~\cite{allan20192017}. We automatically generated binary questions with the structure ``is there [instrument] in this region?" and corresponding masks as rectangular regions with random locations and sizes. Based on the ground-truth label maps, the binary answers were labeled ``yes'' if the region contained at least one pixel of the corresponding instrument and ``no'' otherwise. The questions were balanced to maintain the same amount of ``yes'' and ``no'' answers. 15'580~QA pairs from 1'423 images were used for training, 3'930 from 355 images for validation, and 13'052 from 1'200 images for testing.
    \item[INSEGCAT-VQA.] Frames of cataract surgery videos from the InSegCat~2 dataset~\cite{fox2020insegcat}. We followed the same procedure as in RIS-VQA to generate balanced binary questions with masks and answers.
    The dataset consists of 29'380~QA pairs from 3'519 images for training, 5'306 from 536 images for validation, and 4'322 from 592 images for testing. 
\end{description}
Fig.~\ref{fig:object_distribution} shows the distribution of questions in the three datasets.

\subsection{Baselines and metrics}
\begin{figure}[!t]
\begin{center}
\includegraphics[width=0.93\textwidth]{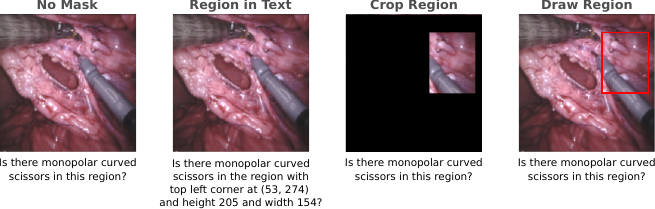}
\caption{Illustration of evaluated baselines for an example image.}
\label{fig:baselines}
\end{center}
\end{figure}

We compare our method to four different baselines, as shown in Fig.~\ref{fig:baselines}:
\begin{description}
    \item[No mask:] no information is provided about the region in the question.
    \item[Region in text~\cite{vu2020question}:] region information is included as text in the question.
    \item[Crop region~\cite{tascon2022consistency}:] image is masked to show only the queried region, with the area outside the region set to zero.
    \item[Draw region:] region is indicated by drawing its boundary on the input image with a distinctive color.
\end{description}
We evaluated the performance of our method using accuracy for the DME-VQA dataset and the area under the receiver operating characteristic (ROC) curve and Average Precision (AP) for the RIS-VQA and INSEGCAT-VQA datasets.

\subsubsection{Implementation details:}

Our VQA architecture uses an LSTM~\cite{hochreiter1997long} with an output dimension~1024 to encode the question and a word embedding size of~300.
We use the ResNet-152~\cite{he2016deep} with ImageNet weights to encode images of size 448$\times$448, generating feature maps with 2048~channels.
In the localized attention block, the visual and textual features are projected into a 512-dimensional space before being combined by element-wise multiplication. The number of glimpses is set to $G=2$ for all experiments.
The classification block is a multi-layer perceptron with a hidden layer of~1024 dimensions. A dropout rate of~0.25 and ReLU activation are used in the localized attention and classifier blocks.

We train our models for 100~epochs using an early stopping condition with patience of 20~epochs. Data augmentation consists of horizontal flips. We use a batch size of 64~samples and the Adam optimizer with a learning rate of $10^{-4}$, which is reduced by a factor of 0.1 when learning stagnates.
Models implemented in PyTorch 1.13.1 and trained on an Nvidia RTX 3090 graphics card.


\subsection{Results} 

\label{sec:results}


\begin{table}[!t]
\begin{center}
\begin{tabular}{llp{0.1cm}lp{0.1cm}lp{0.1cm}lp{0.1cm}c}
\toprule
\multicolumn{1}{c}{\multirow{2}{*}{Method}} & \multicolumn{9}{c}{Accuracy (\%)}                                                                                                                                               \\ \cmidrule{2-10} 
\multicolumn{1}{c}{} & \multicolumn{1}{c}{Overall}      && \multicolumn{1}{c}{Grade}        && \multicolumn{1}{c}{Whole}        && \multicolumn{1}{c}{Macula}       && \multicolumn{1}{c}{Region} \\ \midrule 
No Mask & \multicolumn{1}{c}{61.1 $\pm$ 0.4} && \multicolumn{1}{c}{80.0 $\pm$ 3.7} && \multicolumn{1}{c}{85.7 $\pm$ 1.2} && \multicolumn{1}{c}{84.3 $\pm$ 0.5} && 57.6 $\pm$ 0.4                \\ 
Region in Text~\cite{vu2020question} & \multicolumn{1}{c}{ 60.0 $\pm$ 1.5} && \multicolumn{1}{c}{57.9 $\pm$ 12.5 } && \multicolumn{1}{c}{85.1 $\pm$ 1.9 } && \multicolumn{1}{c}{ 83.2 $\pm$ 2.4} &&     57.7 $\pm$ 1.0        \\ 
Crop Region~\cite{tascon2022consistency}  & \multicolumn{1}{c}{81.4 $\pm$ 0.3} && \multicolumn{1}{c}{78.7 $\pm$ 1.3} && \multicolumn{1}{c}{81.3 $\pm$ 1.7} && \multicolumn{1}{c}{82.3 $\pm$ 1.4} && 81.5 $\pm$ 0.3               \\
Draw Region  & \multicolumn{1}{c}{ 83.0 $\pm$ 1.0} && \multicolumn{1}{c}{79.6 $\pm$ 2.5 } && \multicolumn{1}{c}{77.0 $\pm$ 4.8 } && \multicolumn{1}{c}{\textbf{84.0 $\pm$ 1.9} } &&   83.5 $\pm$ 1.0               \\ 
\ours                                          & \multicolumn{1}{c}{\textbf{84.2 $\pm$ 0.6}} && \multicolumn{1}{c}{\textbf{82.8 $\pm$ 0.4}} && \multicolumn{1}{c}{\textbf{87.0 $\pm$ 1.2}} && \multicolumn{1}{c}{83.0 $\pm$ 1.5} && \textbf{84.2 $\pm$ 0.7}                \\ \bottomrule
\end{tabular}
\end{center}
\caption{Average accuracy for different methods on the DME-VQA dataset. The results shown are the average of 5 models trained with different seeds.}
\label{tab:results_dme}
\end{table}

\begin{table}[!t]
\begin{center}
\begin{tabular}{lp{0.5cm}lp{0.5cm}cp{0.5cm}c}
\toprule
Dataset                       && Method         && AUC && AP \\ \midrule
\multirow{4}{*}{RIS-VQA}      && No Mask    && 0.500 $\pm$ 0.000    &&  0.500 $\pm$ 0.000  \\ 
                                && Region in Text~\cite{vu2020question} &&  0.677 $\pm$ 0.002 && 0.655 $\pm$ 0.003 \\ 
                              && Crop Region~\cite{tascon2022consistency}    &&  0.842 $\pm$ 0.002 && 0.831 $\pm$ 0.002      \\  
                              && Draw Region && 0.835 $\pm$ 0.003 && 0.829 $\pm$ 0.003 \\ 
                              && \ours           && \textbf{0.885 $\pm$ 0.003} && \textbf{0.885 $\pm$ 0.003}\\ \midrule
\multirow{4}{*}{INSEGCAT-VQA} && No Mask    &&   0.500 $\pm$ 0.000  &&  0.500 $\pm$ 0.000   \\  
                              && Region in Text~\cite{vu2020question} &&  0.801 $\pm$ 0.012 && 0.793 $\pm$ 0.014 \\ 
                              && Crop Region~\cite{tascon2022consistency}    &&  0.901 $\pm$ 0.002   && 0.891 $\pm$ 0.003   \\  
                              && Draw Region && 0.910 $\pm$ 0.003 && 0.907 $\pm$ 0.005\\  
                              && \ours           &&  \textbf{0.914 $\pm$ 0.002}   && \textbf{0.915 $\pm$ 0.002}   \\ \bottomrule
\end{tabular}
\end{center}
\caption{Average test AUC and AP for different methods on the RIS-VQA and INSEGCAT-VQA datasets. The results shown are the average over 5~seeds.}
\label{tab:results_ris_insegcat}
\end{table}

\begin{table}[!t]
\ra{1.0}
\begin{center}
\begin{tabular}{p{0.2cm}p{0.13\linewidth}p{0.3cm}LKLKLK}
\toprule
{} &\multirow{2}{*}{Method}  & \multicolumn{7}{c}{Instrument Type} \\
\cmidrule{3-9} 
&         && Large Needle Driver & Monopolar Curved Scissors & Vessel Sealer & Grasping Retractor & Prograsp Forceps & Bipolar Forceps \\ \midrule
&No Mask    && \makecell[tc]{0.500 \\ $\pm$0}              & \makecell[tc]{0.500 \\ $\pm$0}                     & \makecell[tc]{0.500 \\ $\pm$0}            & \makecell[tc]{0.500 \\$\pm$0}                 & \makecell[tc]{0.500 \\ $\pm$0}              & \makecell[tc]{0.500 \\ $\pm$0}              \\ 
&Region in Text~\cite{vu2020question} &&   \makecell[tc]{0.717\\ $\pm$0.003}                  &   \makecell[tc]{0.674\\$\pm$0.001}                        &     \makecell[tc]{0.620\\$\pm$0.011}          &    \makecell[tc]{0.616\\$\pm$0.020}                &    \makecell[tc]{0.647\\$\pm$0.008}              &  \makecell[tc]{0.645\\$\pm$0.003}               \\ 
&Crop Region~\cite{tascon2022consistency}    && \makecell[tc]{0.913 \\$\pm$0.002}               & \makecell[tc]{0.812 \\$\pm$0.003}                     & \makecell[tc]{0.752\\$\pm$0.009}         & \makecell[tc]{0.715\\$\pm$0.015}              & \makecell[tc]{0.773 \\$\pm$0.003}            & \makecell[tc]{0.798 \\$\pm$0.004}           \\ 
&Draw Region &&       \makecell[tc]{0.915\\$\pm$0.003}              &   \makecell[tc]{0.777\\$\pm$0.003}       &   \makecell[tc]{0.783\\$\pm$0.004}      &  \makecell[tc]{0.709\\$\pm$0.012}       &  \makecell[tc]{0.755\\$\pm$0.004}       &     \makecell[tc]{0.805\\$\pm$0.005}            \\ 
&\ours          && \makecell[tc]{\textbf{0.944}\\$\pm$\textbf{0.001}}               & \makecell[tc]{\textbf{0.837} \\$\pm$\textbf{0.005}}                     & \makecell[tc]{\textbf{0.872} \\$\pm$\textbf{0.008}}         & \makecell[tc]{\textbf{0.720} \\$\pm$\textbf{0.031}}              & \makecell[tc]{\textbf{0.834} \\$\pm$\textbf{0.006}}            & \makecell[tc]{\textbf{0.880} \\$\pm$\textbf{0.003}}           
\\ \bottomrule
\end{tabular}
\end{center}
\caption{Average test AUC for different methods on the RIS-VQA dataset as a function of instrument type. Results are averaged over 5 models trained with different seeds. The corresponding table for INSEGCAT-VQA is available in the Supplementary Materials.
}
\label{tab:results_ris_object}
\end{table}

Our method outperformed all considered baselines on the DME-VQA (Table~\ref{tab:results_dme}), the RIS-VQA, and the INSEGCAT-VQA datasets (Table~\ref{tab:results_ris_insegcat}), highlighting the importance of contextual information in answering localized questions. Context proved to be particularly critical in distinguishing between objects of similar appearance, such as the bipolar and prograsp forceps in RIS-VQA, where our method led to an 8~percent point performance improvement (Table~\ref{tab:results_ris_object}). In contrast, the importance of context was reduced when dealing with visually distinct objects, resulting in smaller performance gains as observed in the INSEGCAT-VQA dataset. For example, despite not incorporating contextual information, the baseline \emph{crop region} still benefited from correlations between the location of the region and the instrument mentioned in the question (\eg,~the eye retractor typically appears at the top or the bottom of the image), enabling it to achieve competitive performance levels that are less than 2~percent points lower than our method~(Table~\ref{tab:results_ris_insegcat}, bottom).

\begin{figure}[!t]
\begin{center}
\includegraphics[width=\textwidth]{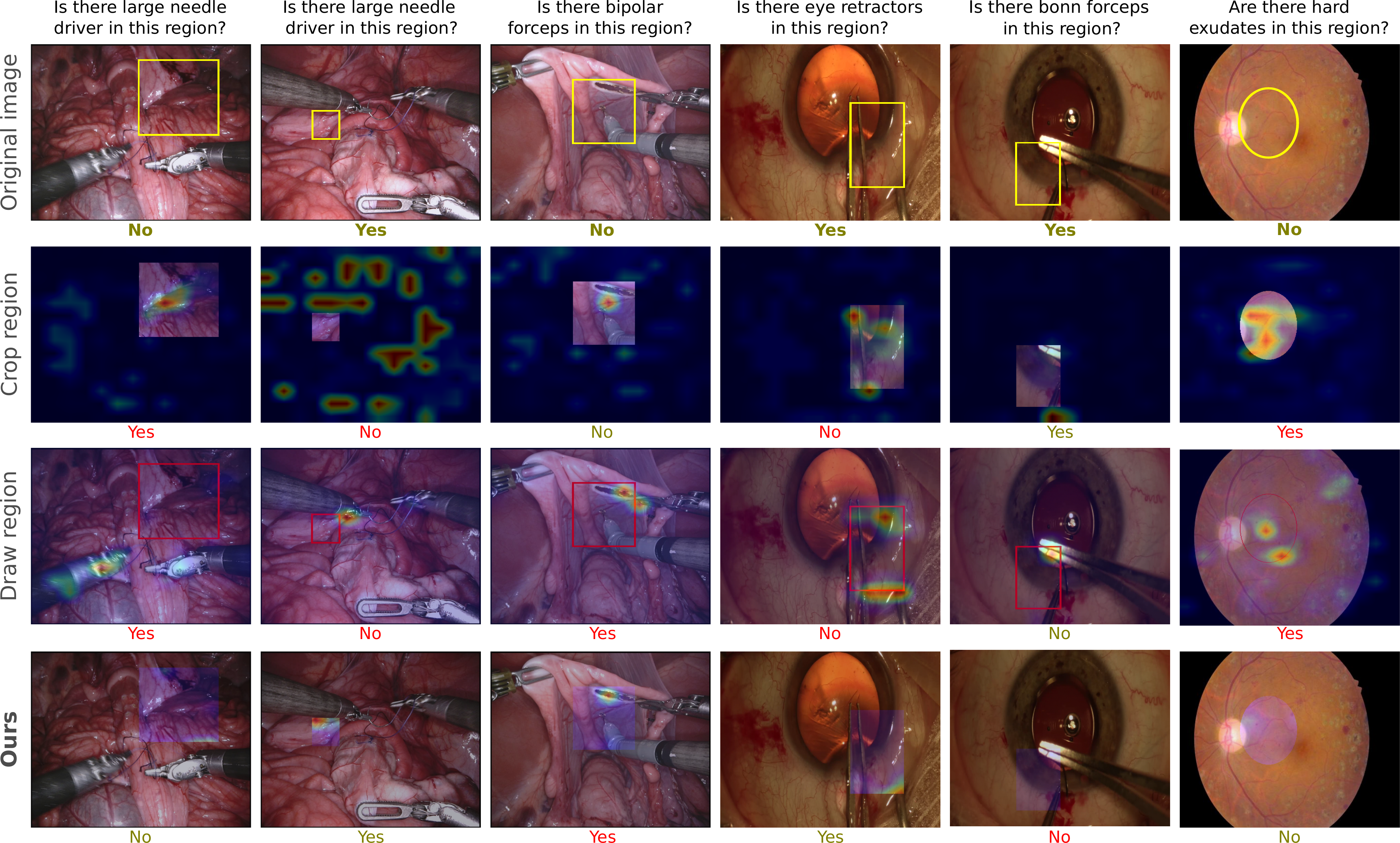}
\caption{Qualitative examples on the RIS-VQA dataset (columns~1-3), INSEGCAT-VQA (columns~4-5), and DME-VQA (last column). Only the strongest baselines were considered in this comparison. The first row shows the image, the region, and the ground truth answer. Other rows show the overlaid attention maps and the answers produced by each model. Wrong answers are shown in red. Additional examples are available in the Supplementary Materials.}
\label{fig:examples_att}
\end{center}
\end{figure}

Similar to our method, the baseline \emph{draw region} incorporates contextual information when answering localized questions. However, we observed that drawing regions on the image can interfere with the computation of guided attention maps, leading to incorrect predictions (Fig.~\ref{fig:examples_att}, column~4). In addition, the lack of masking of the attention maps often led the model to wrongly consider areas beyond the region of interest while answering questions~(Fig.~\ref{fig:examples_att}, column~1). 

When analyzing mistakes made by our model, we observe that they tend to occur when objects or background structures in the image look similar to the object mentioned in the question~(Fig.~\ref{fig:examples_att}, column~3). Similarly, false predictions were observed when only a few pixels of the object mentioned in the question were present in the region.



\section{Conclusions}
\label{sec:conc}

In this paper, we proposed a novel VQA architecture to answer questions about regions. We compare the performance of our approach against several baselines and across three different datasets. By focusing the model's attention on the region after considering the evidence in the full image, we show how our method brings improvements, especially when the complete image context is required to answer the questions. Future works include studying the agreement between answers to questions about concentric regions, as well as the agreement between questions about images and regions.

\clearpage 
\bibliographystyle{splncs04}
\bibliography{mybibliography}

\begin{thebibliography}{10}
\providecommand{\url}[1]{\texttt{#1}}
\providecommand{\urlprefix}{URL }
\providecommand{\doi}[1]{https://doi.org/#1}

\bibitem{allan20192017}
Allan, M., Shvets, A., Kurmann, T., Zhang, Z., Duggal, R., Su, Y.H., Rieke, N.,
  Laina, I., Kalavakonda, N., Bodenstedt, S., et~al.: 2017 robotic instrument
  segmentation challenge. arXiv preprint arXiv:1902.06426  (2019)

\bibitem{antol2015vqa}
Antol, S., Agrawal, A., Lu, J., Mitchell, M., Batra, D., Zitnick, C.L., Parikh,
  D.: Vqa: Visual question answering. In: Proceedings of the IEEE international
  conference on computer vision. pp. 2425--2433 (2015)

\bibitem{ben2017mutan}
Ben-Younes, H., Cadene, R., Cord, M., Thome, N.: Mutan: Multimodal tucker
  fusion for visual question answering. In: Proceedings of the IEEE
  international conference on computer vision. pp. 2612--2620 (2017)

\bibitem{do2021multiple}
Do, T., Nguyen, B.X., Tjiputra, E., Tran, M., Tran, Q.D., Nguyen, A.: Multiple
  meta-model quantifying for medical visual question answering. In: Medical
  Image Computing and Computer Assisted Intervention--MICCAI 2021: 24th
  International Conference, Strasbourg, France, September 27--October 1, 2021,
  Proceedings, Part V 24. pp. 64--74. Springer (2021)

\bibitem{fox2020insegcat}
Fox, M., Taschwer, M., Schoeffmann, K.: Pixel-based tool segmentation in
  cataract surgery videos with mask {R-CNN}. In: de~Herrera, A.G.S.,
  Gonz{\'{a}}lez, A.R., Santosh, K.C., Temesgen, Z., Kane, B., Soda, P. (eds.)
  33rd {IEEE} International Symposium on Computer-Based Medical Systems, {CBMS}
  2020, Rochester, MN, USA, July 28-30, 2020. pp. 565--568. {IEEE} (2020).
  \doi{10.1109/CBMS49503.2020.00112},
  \url{https://doi.org/10.1109/CBMS49503.2020.00112}

\bibitem{gong2021cross}
Gong, H., Chen, G., Liu, S., Yu, Y., Li, G.: Cross-modal self-attention with
  multi-task pre-training for medical visual question answering. In:
  Proceedings of the 2021 International Conference on Multimedia Retrieval. pp.
  456--460 (2021)

\bibitem{goyal2017making}
Goyal, Y., Khot, T., Summers-Stay, D., Batra, D., Parikh, D.: Making the v in
  vqa matter: Elevating the role of image understanding in visual question
  answering. In: Proceedings of the IEEE Conference on Computer Vision and
  Pattern Recognition. pp. 6904--6913 (2017)

\bibitem{ImageCLEFVQA_Med2018}
Hasan, S.A., Ling, Y., Farri, O., Liu, J., Lungren, M., M\"uller, H.: Overview
  of the {ImageCLEF} 2018 medical domain visual question answering task. In:
  CLEF2018 Working Notes. {CEUR} Workshop Proceedings, CEUR-WS.org
  $<$http://ceur-ws.org$>$, Avignon, France (September 10-14 2018)

\bibitem{he2016deep}
He, K., Zhang, X., Ren, S., Sun, J.: Deep residual learning for image
  recognition. In: Proceedings of the IEEE conference on computer vision and
  pattern recognition. pp. 770--778 (2016)

\bibitem{hochreiter1997long}
Hochreiter, S., Schmidhuber, J.: Long short-term memory. Neural computation
  \textbf{9}(8),  1735--1780 (1997)

\bibitem{hudson2019gqa}
Hudson, D.A., Manning, C.D.: Gqa: a new dataset for compositional question
  answering over real-world images. arXiv preprint arXiv:1902.09506
  \textbf{3}(8) (2019)

\bibitem{kim2016hadamard}
Kim, J.H., On, K.W., Lim, W., Kim, J., Ha, J.W., Zhang, B.T.: Hadamard product
  for low-rank bilinear pooling. arXiv preprint arXiv:1610.04325  (2016)

\bibitem{liao2020aiml}
Liao, Z., Wu, Q., Shen, C., Van Den~Hengel, A., Verjans, J.: Aiml at vqa-med
  2020: Knowledge inference via a skeleton-based sentence mapping approach for
  medical domain visual question answering  (2020)

\bibitem{liu2021slake}
Liu, B., Zhan, L.M., Xu, L., Ma, L., Yang, Y., Wu, X.M.: Slake: A
  semantically-labeled knowledge-enhanced dataset for medical visual question
  answering. In: 2021 IEEE 18th International Symposium on Biomedical Imaging
  (ISBI). pp. 1650--1654. IEEE (2021)

\bibitem{liu2019effective}
Liu, F., Peng, Y., Rosen, M.P.: An effective deep transfer learning and
  information fusion framework for medical visual question answering. In:
  International Conference of the Cross-Language Evaluation Forum for European
  Languages. pp. 238--247. Springer (2019)

\bibitem{mani2020point}
Mani, A., Yoo, N., Hinthorn, W., Russakovsky, O.: Point and ask: Incorporating
  pointing into visual question answering. arXiv preprint arXiv:2011.13681
  (2020)

\bibitem{Nguyen19}
Nguyen, B.D., Do, T.T., Nguyen, B.X., Do, T., Tjiputra, E., Tran, Q.D.:
  Overcoming data limitation in medical visual question answering. In: Shen,
  D., Liu, T., Peters, T.M., Staib, L.H., Essert, C., Zhou, S., Yap, P.T.,
  Khan, A. (eds.) Medical Image Computing and Computer Assisted Intervention --
  MICCAI 2019. pp. 522--530. Springer International Publishing, Cham (2019)

\bibitem{pelka2018radiology}
Pelka, O., Koitka, S., R{\"u}ckert, J., Nensa, F., Friedrich, C.M.: Radiology
  objects in context (roco): a multimodal image dataset. In: Intravascular
  Imaging and Computer Assisted Stenting and Large-Scale Annotation of
  Biomedical Data and Expert Label Synthesis: 7th Joint International Workshop,
  CVII-STENT 2018 and Third International Workshop, LABELS 2018, Held in
  Conjunction with MICCAI 2018, Granada, Spain, September 16, 2018, Proceedings
  3. pp. 180--189. Springer (2018)

\bibitem{ren2020cgmvqa}
Ren, F., Zhou, Y.: Cgmvqa: A new classification and generative model for
  medical visual question answering. IEEE Access  \textbf{8},  50626--50636
  (2020)

\bibitem{tan2019lxmert}
Tan, H., Bansal, M.: Lxmert: Learning cross-modality encoder representations
  from transformers. arXiv preprint arXiv:1908.07490  (2019)

\bibitem{tascon2022consistency}
Tascon-Morales, S., M{\'a}rquez-Neila, P., Sznitman, R.: Consistency-preserving
  visual question answering in medical imaging. In: Medical Image Computing and
  Computer Assisted Intervention--MICCAI 2022: 25th International Conference,
  Singapore, September 18--22, 2022, Proceedings, Part VIII. pp. 386--395.
  Springer (2022)

\bibitem{vu2020question}
Vu, M.H., L{\"o}fstedt, T., Nyholm, T., Sznitman, R.: A question-centric model
  for visual question answering in medical imaging. IEEE transactions on
  medical imaging  \textbf{39}(9),  2856--2868 (2020)

\bibitem{yu2023question}
Yu, Y., Li, H., Shi, H., Li, L., Xiao, J.: Question-guided feature pyramid
  network for medical visual question answering. Expert Systems with
  Applications  \textbf{214},  119148 (2023)

\bibitem{zhan2020medical}
Zhan, L.M., Liu, B., Fan, L., Chen, J., Wu, X.M.: Medical visual question
  answering via conditional reasoning. In: Proceedings of the 28th ACM
  International Conference on Multimedia. pp. 2345--2354 (2020)

\end{thebibliography}

\end{document}